\documentclass[conference]{IEEEtran}
\IEEEoverridecommandlockouts
\usepackage{cite}
\usepackage{amsmath,amssymb,amsfonts}
\usepackage{algorithmic}
\usepackage{graphicx}
\usepackage{textcomp}
\usepackage{xcolor}
\usepackage{booktabs}
\usepackage{fontspec}
\usepackage{graphicx}
\usepackage{caption}
\usepackage{subcaption}
\usepackage{float}
\usepackage{makecell}
\newfontfamily\bengalifont[
Path = ./,
UprightFont = NotoSansBengali-Regular.ttf,
BoldFont = NotoSansBengali-Bold.ttf
]{NotoSansBengali}
\newcommand{\textbn}[1]{{\bengalifont #1}}

\def\BibTeX{{\rm B\kern-.05em{\sc i\kern-.025em b}\kern-.08em
    T\kern-.1667em\lower.7ex\hbox{E}\kern-.125emX}}
\begin{document}

\title{A Robust Deep Learning Framework for Bangla License Plate Recognition Using YOLO and Vision–Language OCR\\

}

\author{\IEEEauthorblockN{Nayeb Hasin}
\IEEEauthorblockA{\textit{Department of Electrical and}\\
\textit{Electronic Engineering} \\
\textit{Islamic University of Technology}\\
Gazipur, Bangladesh \\
nayebhasin@iut-dhaka.edu}
\and
\IEEEauthorblockN{Md. Arafath Rahman Nishat}
\IEEEauthorblockA{\textit{Department of Mechanical Engineering} \\
\textit{Bangladesh University Of}\\
\textit{Engineering And Technology}\\
Dhaka, Bangladesh \\
arnishat100@gmail.com}
\and
\IEEEauthorblockN{Mainul Islam}
\IEEEauthorblockA{\textit{Department of Electrical and}\\
\textit{Electronic Engineering} \\
\textit{Islamic University of Technology}\\
Gazipur, Bangladesh \\
mainulislam@iut-dhaka.edu}
\and \and
\IEEEauthorblockN{Khandakar Shakib Al Hasan}
\IEEEauthorblockA{\textit{Department of Electrical and Electronic Engineering} \\
\textit{Islamic University of Technology}\\
Gazipur, Bangladesh  \\
shakibalhasan@iut-dhaka.edu}
\and
\IEEEauthorblockN{Asif Newaz}
\IEEEauthorblockA{\textit{Department of Electrical and Electronic Engineering} \\
\textit{Islamic University of Technology}\\
Gazipur, Bangladesh  \\
eee.asifnewaz@iut-dhaka.edu}
}

\maketitle

\begin{abstract}
An Automatic License Plate Recognition (ALPR) system constitutes a crucial element in an intelligent traffic management system. However, the detection of Bangla license plates remains challenging because of the complicated character scheme and uneven layouts. 
This paper presents a robust Bangla License Plate Recognition system that integrates a deep learning–based object detection model for license plate localization with Optical Character Recognition (OCR) for text extraction. Multiple object detection architectures, including U-Net and several YOLO (You Only Look Once) variants, are compared for license plate localization. A key contribution includes a novel two-stage adaptive training strategy for YOLOv8, incorporating phase-aware augmentation and progressive layer unfreezing to improve robustness under real-world variations. Extensive experiments show that the proposed approach achieves 97.83\% accuracy and 91.3\% IoU, outperforming multiple YOLO variants. Additionally, a VisionEncoderDecoder-based OCR framework with BanglaBERT achieves superior character-level performance (CER 0.1323). The proposed system also shows a consistent performance when tested on an external dataset that has been curated for this study purpose, highlighting its practical applicability. The dataset offers completely different environment and lighting conditions compared to the training sample. Overall, our proposed system provides a robust and reliable solution for Bangla license plate recognition and performs effectively across diverse real-world scenarios, including variations in lighting, noise, and plate styles. These strengths make it well suited for deployment in intelligent transportation applications such as automated law enforcement and access control. 

\end{abstract}

\section{Introduction}
All the registered vehicles are expected to have a license plate. A license plate is a unique alphanumeric identification mark that is issued to every registered automobile that is used as a legal identification and control mark. It allows the government to monitor automobile operations including ownership, taxation, law enforcement and traffic control. Over the last several years, Automatic License Plate Recognition (ALPR) systems have grown to be an essential part of modern intelligent transportation systems providing automated tracking of vehicles, tolls, and access control without human involvement.

ALPR technologies have attained considerable accuracy for various widely used scripts, such as English, Latin and Arabic \cite{Laroca_2018}\cite{hussain2020developing}\cite{Moussaoui2023}. However, the recognition of Bangla license plates presents considerable difficulties. It is due to the complex structure of Bangla characters, the presence of compound letters, diverse fonts, and inconsistent plate designs. Also, things like poor lighting, motion blur, and occlusions can make images look worse. It makes it even harder to do tasks like segmentation and recognition.

In order to improve law enforcement and automation of traffic in Bangladesh and other places where the Bengali language is spoken, it is necessary to create a powerful system of Bangla License Plate and Number Recognition (BLPNR). The aim of this research is to design and test an automated recognition system that has the potential to precisely detect, segment and distinguish Bangla license plates under various real life scenarios.

We present a robust, two-stage deep learning (DL) methodology to overcome the challenges posed by the Bangla script and adverse imaging conditions: \textbf{License Plate Localization} and \textbf{Text Recognition}. The initial stage, License Plate Localization, is responsible for accurately identifying the license plate's bounding box within the input image. To determine the most effective solution, we conducted a rigorous comparative analysis of multiple object detection architectures, including U-Net and several YOLO (You Only Look Once) variants such as YOLOv5m, YOLOv7m, YOLOv8m, YOLOv9m, and YOLOv11m.

The core of our proposed approach is a novel two-stage adaptive training strategy specifically built upon the YOLOv8 architecture. This strategy incorporates aggressive, phase-aware data augmentation and dynamic layer unfreezing to systematically improve the model's resilience to variations in viewpoint, lighting, and noise. This specialized model architecture and training regimen was demonstrated to achieve superior performance, successfully handling the uneven layouts common in Bangla plates. The second stage performs Optical Character Recognition (OCR) on the cropped license plate image produced by the localization module. The problem is formulated as a sequence generation task and solved using a Vision-Encoder-Decoder (VED) architecture, which includes BanglaBERT tokenizer for the OCR model.

Another key contribution of this work is the curation of a dedicated dataset for external validation under diverse real-world conditions. While the models are trained on publicly available datasets, external validation is conducted on the curated dataset to assess the effectiveness of the proposed architecture. The dataset includes license plates from all districts, encompassing variations in lighting, weather conditions, and camera orientations, thereby providing a wide range of real-world scenarios.

\section{Literature Review}
In the past twenty years, Automatic License Plate Recognition (ALPR) has progressed significantly, evolving from traditional image-processing approaches to DL-based end-to-end systems. The original ALPR methods were based on manually designed features and edge detection for localization and template matching for recognition of characters. However, the performance of these methods was very sensitive to variations in lighting, orientation, and license plate design. Recent progress in DL, in particular convolutional neural networks (CNNs), have greatly improved the detection and recognition accuracy in diverse conditions.

Early ALPR methods relied on manually designed features, edge detection for localization, and template matching for character recognition, but these were highly sensitive to variations in lighting, orientation, and plate design. Recent advancements in deep learning, particularly convolutional neural networks (CNNs) and state-of-the-art object detectors like YOLO, have greatly improved detection and recognition accuracy under diverse conditions \cite{laroca2018robust, Meesad2025}. Laroca et al. \cite{Laroca_2021} developed a layout-independent ALPR framework using the YOLO object detector, achieving over 96\% end-to-end accuracy across various datasets. Researchers have also successfully integrated YOLOv7 with image processing and OCR post-processing to enhance detection and recognition, particularly for mixed-script environments \cite{Moussaoui2023}. However, many high-performing systems primarily target Latin-script plates, where character structures and layouts are simpler. This often leads to degraded performance when these systems are applied directly to complex non-Latin scripts.

For Arabic-script plates, challenges such as right-to-left directionality, complex ligatures, and dual-language layouts significantly complicate recognition. Hussain and Hathal \cite{hussain2020developing} proposed an Arabic plate recognition system utilizing Canny edge detection and an artificial neural network. While effective, its reliance on traditional edge detection limits its robustness to real-world variations like blur and inconsistent lighting. To address these limitations, modern intelligent traffic systems have shifted toward custom deep learning frameworks specifically tailored to handle the intricacies of Arabic license plates \cite{Sayedelahl_2024}. While the Arabic script shares structural complexities with Bangla (e.g., ligatures and non-linear character placement), existing solutions are heavily tuned to regional formats, reducing their direct transferability to Bangladeshi BRTA plates.

Research specifically targeting Bangla (Bengali) license plates remains relatively recent. Early computer vision efforts, such as the contour analysis and prediction algorithm proposed by Pervej et al. \cite{pervej_realtime_bangla}, demonstrated real-time capabilities but were often sensitive to environmental noise and plate degradation. The introduction of comprehensive multi-step deep learning models by Shomee and Sams \cite{9647284} marked a significant step forward, utilizing YOLO-based detection alongside generative models to recognize all types of Bangladeshi vehicles. To further address image clarity and low-resolution constraints, Afrin et al. \cite{afrin2023bengalilicenseplaterecognition} paired CNNs with GFP-GAN for image restoration. Dedicated architectures have also emerged to tackle the unique script; Onim et al. \cite{onim2022blpnetnewdnnmodel} introduced BLPnet, which integrates a new DNN model with a Bangla-specific OCR engine to maximize character recognition accuracy. Most recently, researchers have begun leveraging the latest detector generations, with Ismail and Ahamed \cite{Ismail2025} applying YOLOv8 to specifically tackle the localization and recognition of Bangladeshi vehicle plates.

The literature shows that ALPR for Bangla license plate recognition continues to pose a significant challenge due to restricted datasets, intricate character structures, and variable plate configurations. Existing DL approaches for Bangla often lack a comparative study of the latest, most robust YOLO models and an adaptive training strategy specifically designed to maximize localization IoU under diverse real-world constraints, which is the key focus of the proposed methodology.

\section{Dataset}
\label{Dataset}
The Bangladesh Road Transport Authority (BRTA) is the regulatory agency that is responsible for supervising, managing, and ensuring discipline and safety within the country’s road transport sector. In 2012, the agency introduced a new Retro Reflective License Plate, widely known as the digital license plate. Since then, it is mandated by law that every vehicle  display this license plate on its rear. 

All text on the license plate should be in the Bangla language. The first word of the top row indicates the district where the vehicle is registered. The second word shows whether the vehicle is registered in a metropolitan zone. The last character in the top row is for the category of the vehicle. The bottom row consists of six digits. The first two digits indicate the vehicle’s class registration number. The remaining four digits are the vehicle’s serial number. Two sample license plates are illustrated in Figure \ref{fig:licenseplate} below.

\begin{figure}[t]
    \centering
    \begin{subfigure}[b]{0.48\linewidth}
        \centering
        \includegraphics[width=\linewidth]{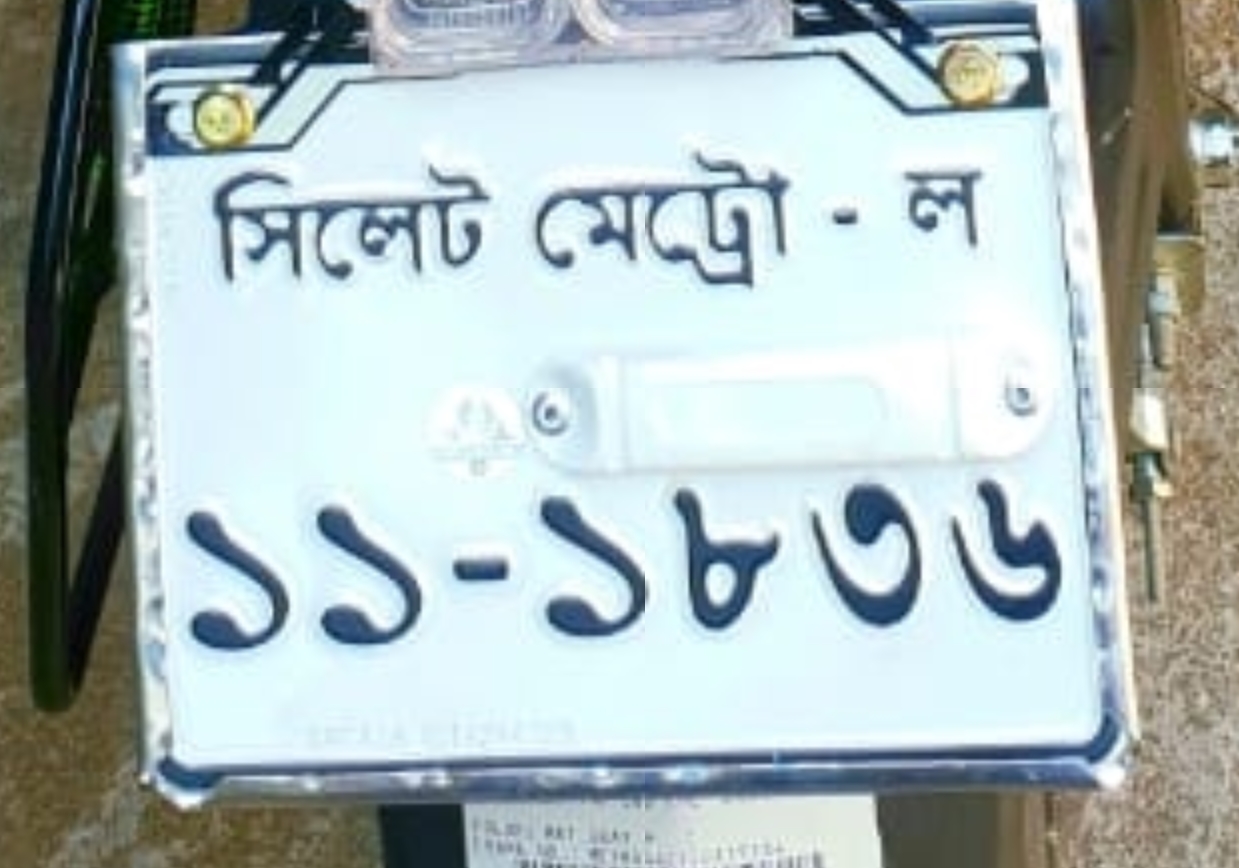}
        \caption{ Private Vehicle}
        \label{fig:image1}
    \end{subfigure}
    \hfill
    \begin{subfigure}[b]{0.48\linewidth}
        \centering
        \includegraphics[width=\linewidth]{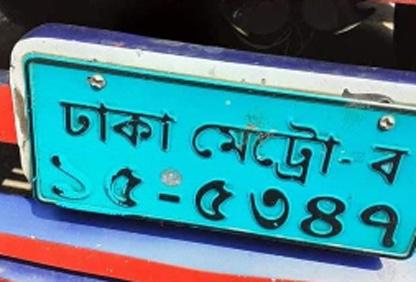}
        \caption{Trading Vehicle}
        \label{fig:image2}
    \end{subfigure}

    \caption{BRTA Standard Vehicle Registration Plate}
    \label{fig:licenseplate}
\end{figure}

\begin{figure}[b]
    \centering
    \begin{subfigure}[b]{0.48\linewidth}
        \centering
        \includegraphics[width=\linewidth]{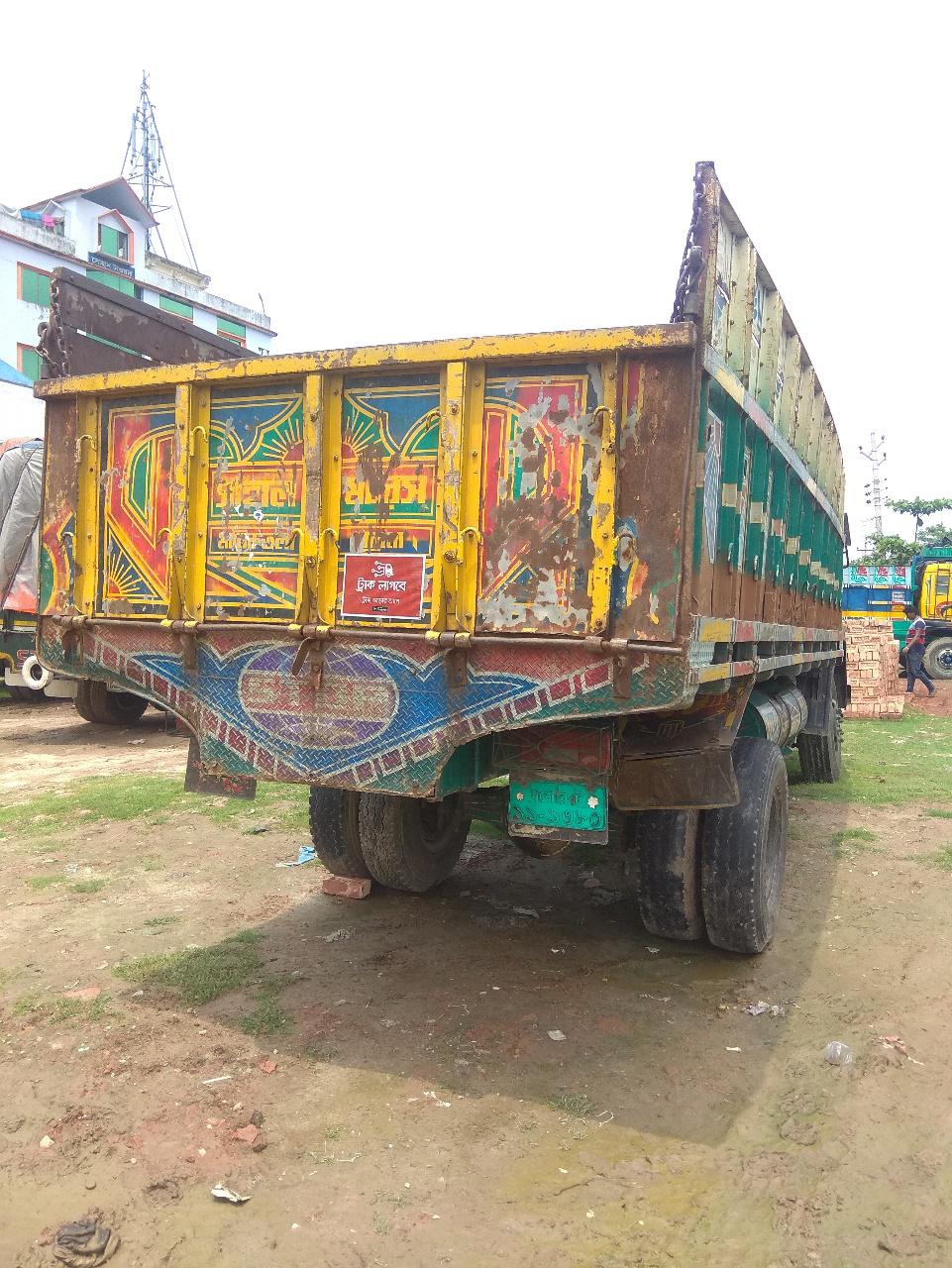}
        \label{fig:image1}
    \end{subfigure}
    \hfill
    \begin{subfigure}[b]{0.48\linewidth}
        \centering
        \includegraphics[width=\linewidth]{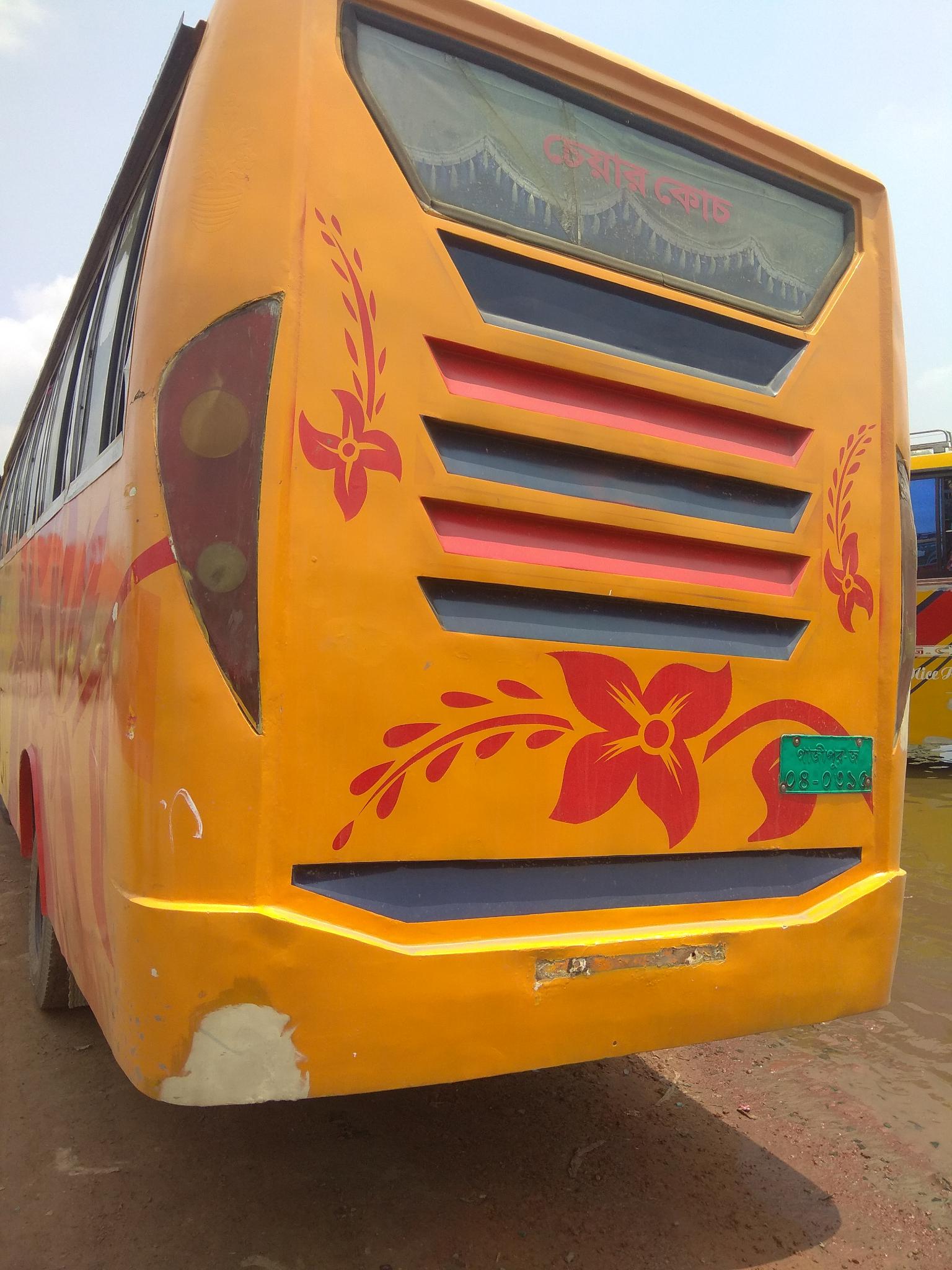}
        \label{fig:image2}
    \end{subfigure}

    \caption{Images for License Plate Detection}
    \label{fig:localizationdataset}
\end{figure}

This study utilizes two separate datasets: one for license plate localization and another for text extraction. For localization, “Bangladeshi License Plate Recognition Dataset” by Syed Nahin Hossain et al. is used \cite{syed_nahin_hossain2023}. This dataset comprises 6,517 images with and without license plates. The images are divided into three groups: 5,687 for training, 277 images for validation during the training process, and 553 for testing the model's performance after training. The images vary in camera angle and contain different levels of noise. The dataset also provides the annotation files in Pascal VOC format, which can be readily employed to train the localization model with YOLO. For training the localization model with the U-Net, an individual pixel mask was generated for each image based on the annotation files. Two sample images from the dataset are shown in Figure \ref{fig:localizationdataset}.

\begin{figure}[t]
    \centering
    \begin{subfigure}[b]{0.48\linewidth}
        \centering
        \includegraphics[width=\linewidth]{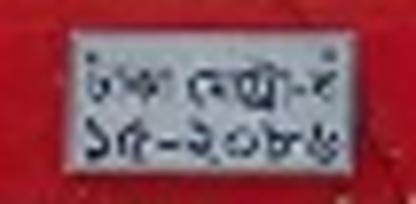}
        \caption{Real}
        \label{fig:image1}
    \end{subfigure}
    \hfill
    \begin{subfigure}[b]{0.48\linewidth}
        \centering
        \includegraphics[width=\linewidth]{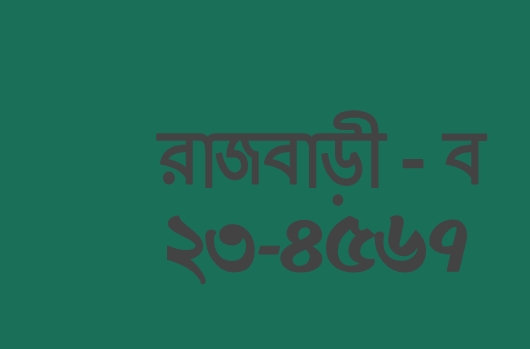}
        \caption{Synthetic}
        \label{fig:image2}
    \end{subfigure}

    \caption{Images for Text Extraction}
    \label{fig:textdataset}
\end{figure}

Our main localization dataset contains license plate images collected mostly from Dhaka city. However, training only on Dhaka-centric data can make the OCR model too specific to that environment. To make the OCR model more general and reliable in different regions, we also used an open-source dataset called “Bangla License Plate Dataset with Annotations” by MMHQ\cite{mmhq}. The dataset covers license plates from 64 distinct districts of Bangladesh. It is a collection of 13629 cropped images of license plates with different levels of noise and blurriness. The training group consists of 12680 images, with approximately 26\% synthetic images. The validation and test directories have 564 and 385 images, respectively. The images are annotated with text labels for 102 classes. The classes include digits \textbn{০-৯}, the word \textbn{"মেট্রো"}, and different Bengali alphabet characters. Two sample images from the dataset is shown in Figure \ref{fig:textdataset}.

An additional dataset, prepared by our team, is used for external validation. This is used to evaluate the model's robustness and reliability in real-world scenarios. The dataset comprises 276 test images captured from the CCTV camera of a toll booth in low lighting conditions. The images were annotated in YOLOv8 format using Roboflow. The sample images from the dataset are shown in Figure \ref{fig:customdataset} below.
 \begin{figure}[b]
    \centering
    \begin{subfigure}[b]{0.48\linewidth}
        \centering
        \includegraphics[width=\linewidth]{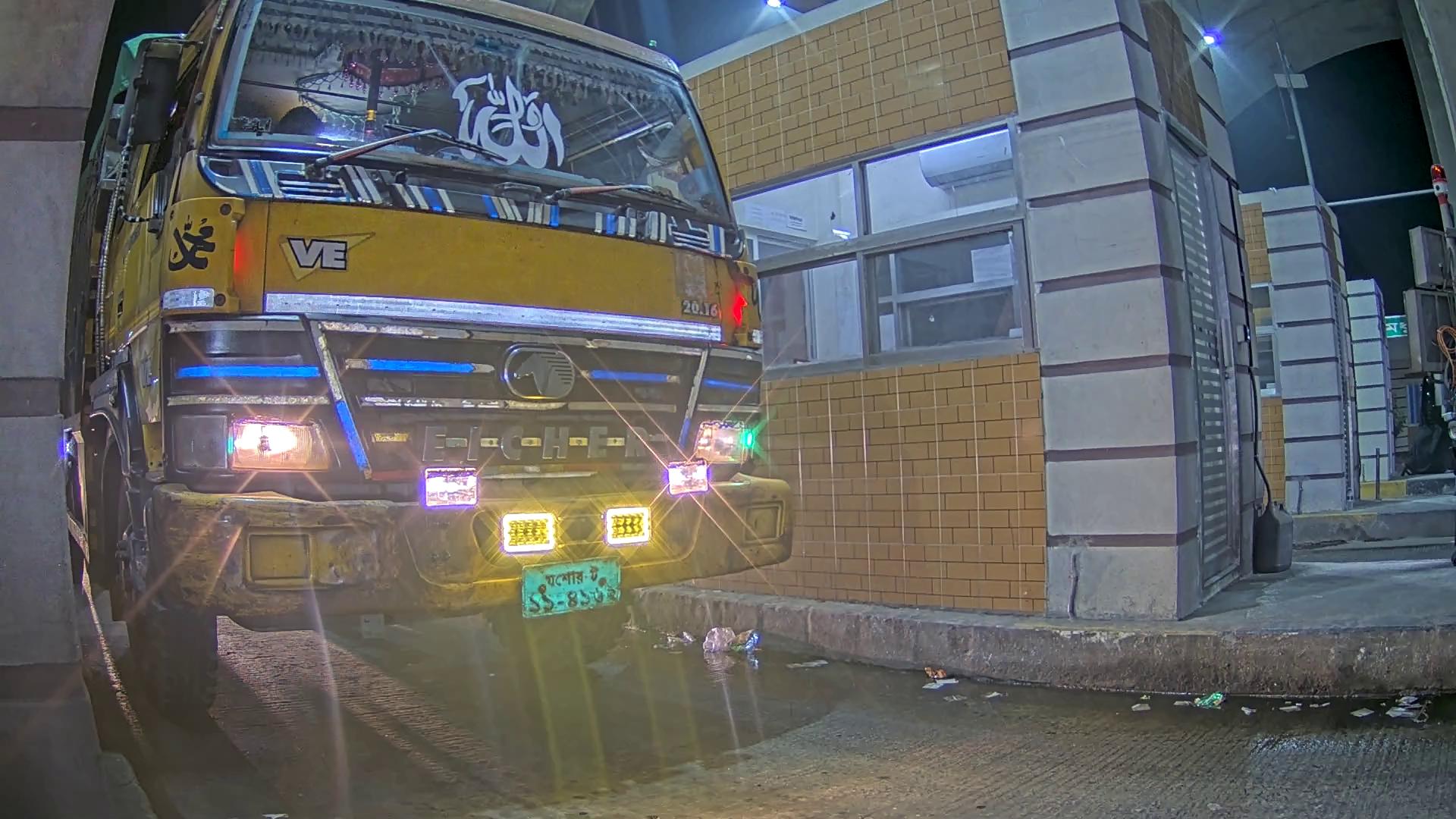}
        \label{fig:image1}
    \end{subfigure}
    \hfill
    \begin{subfigure}[b]{0.48\linewidth}
        \centering
        \includegraphics[width=\linewidth]{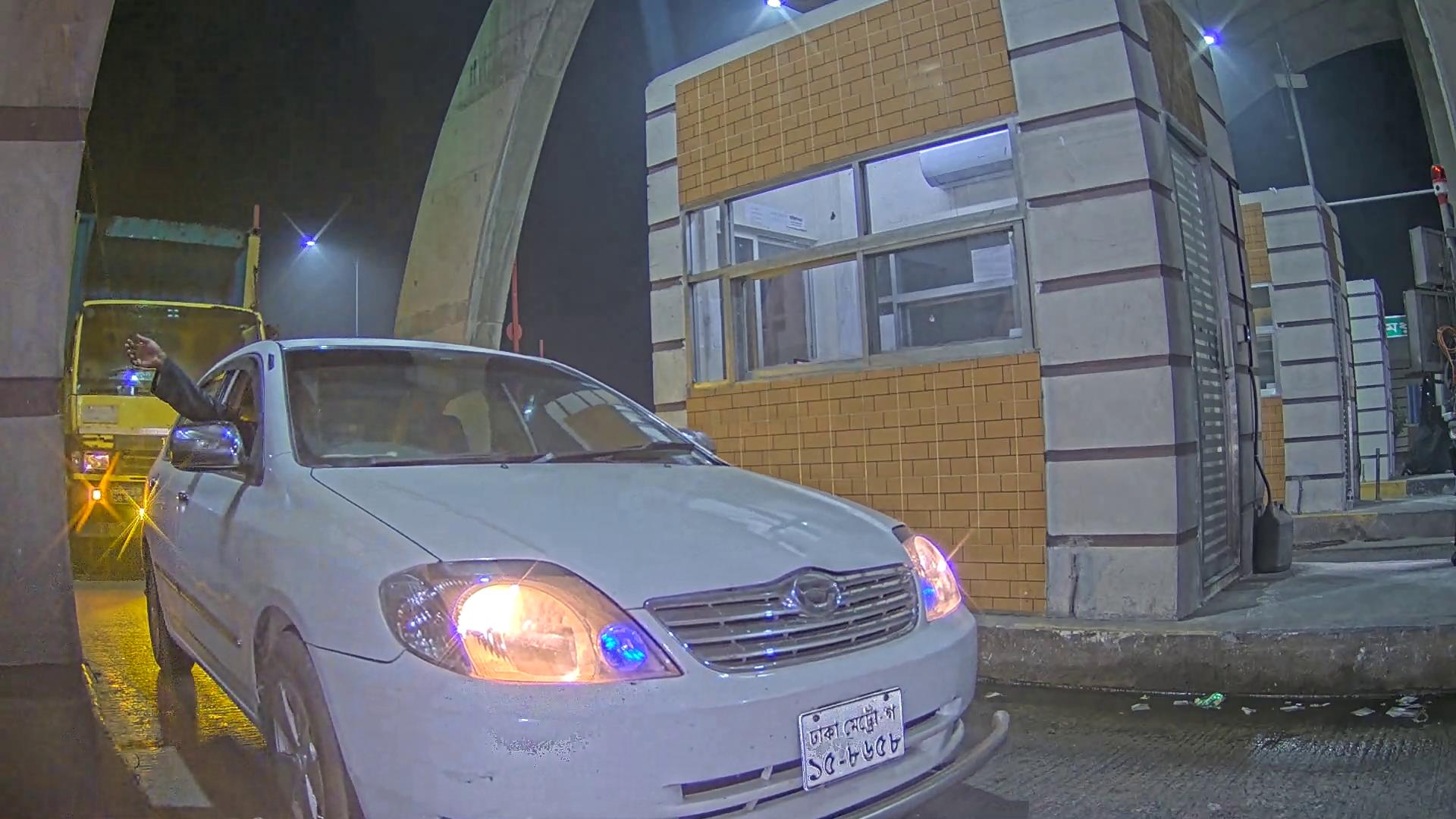}
        \label{fig:image2}
    \end{subfigure}

    \caption{Images for External Validation}
    \label{fig:customdataset}
\end{figure}

\section{Methodology for License Plate Detection}
Though numerous image processing-based methods have been employed for detecting license plates, their performance often degrades under specific circumstances. This study aims to develop a more robust solution that remains reliable under a wide range of environmental conditions, including various lighting levels, weather variations, and changes in camera orientation.

In this study, we experimented with several DL–based detection frameworks, including U-Net and multiple variants of the YOLO architecture, including v5m, v7m, v8m, v9m, and v11m. After evaluating the performance of these models, YOLOv8m was chosen due to its superior detection capability, reflected by its improved IoU. The medium-sized models were intentionally selected to ensure satisfactory accuracy with faster processing and reduced hardware requirements.

The YOLOv8m architecture is then combined with a two-stage adaptive training strategy and phase-aware data augmentation to improve detection performance. In this approach, the model is trained in two stages with a batch size of 10 images.

The first training stage focused on aggressive feature learning over 35 epochs with high learning rates, aggressive momentum, and weight decay. Progressive layer unfreezing was implemented in this stage. This froze 12 layers initially, 8 layers during the middle phase, and 4 layers towards the end. Among different progressive unfreezing schedules, this configuration was selected as it provided the best balance between feature stability and the highest validation performance. The loss weighting strategy combined box loss, class loss and DFL loss. Augmentation in this stage was spatially intensive, with ±8° rotation, 15\% translation, 0.7× scaling, and ±3° shear. To preserve critical text properties, perspective distortion and vertical flipping were avoided. Advanced composite augmentations include 100\% mosaic probability (4-image composites), 15\% mix-up (image–label blending), 40\% copy-paste (object transplantation), and 50\% horizontal flipping were also applied. Photometric augmentations were moderate, with ±1\% hue, ±80\% saturation, and ±50\% brightness variations. The convergence of the training process was monitored using a sliding window of 8 epochs. It calculated the improvements in mAP values and triggered at a threshold of 0.001. The early stopping patience of this stage was 15 epochs.

\begin{figure}[t]
    \centering
    \includegraphics[width=0.9\columnwidth,keepaspectratio]{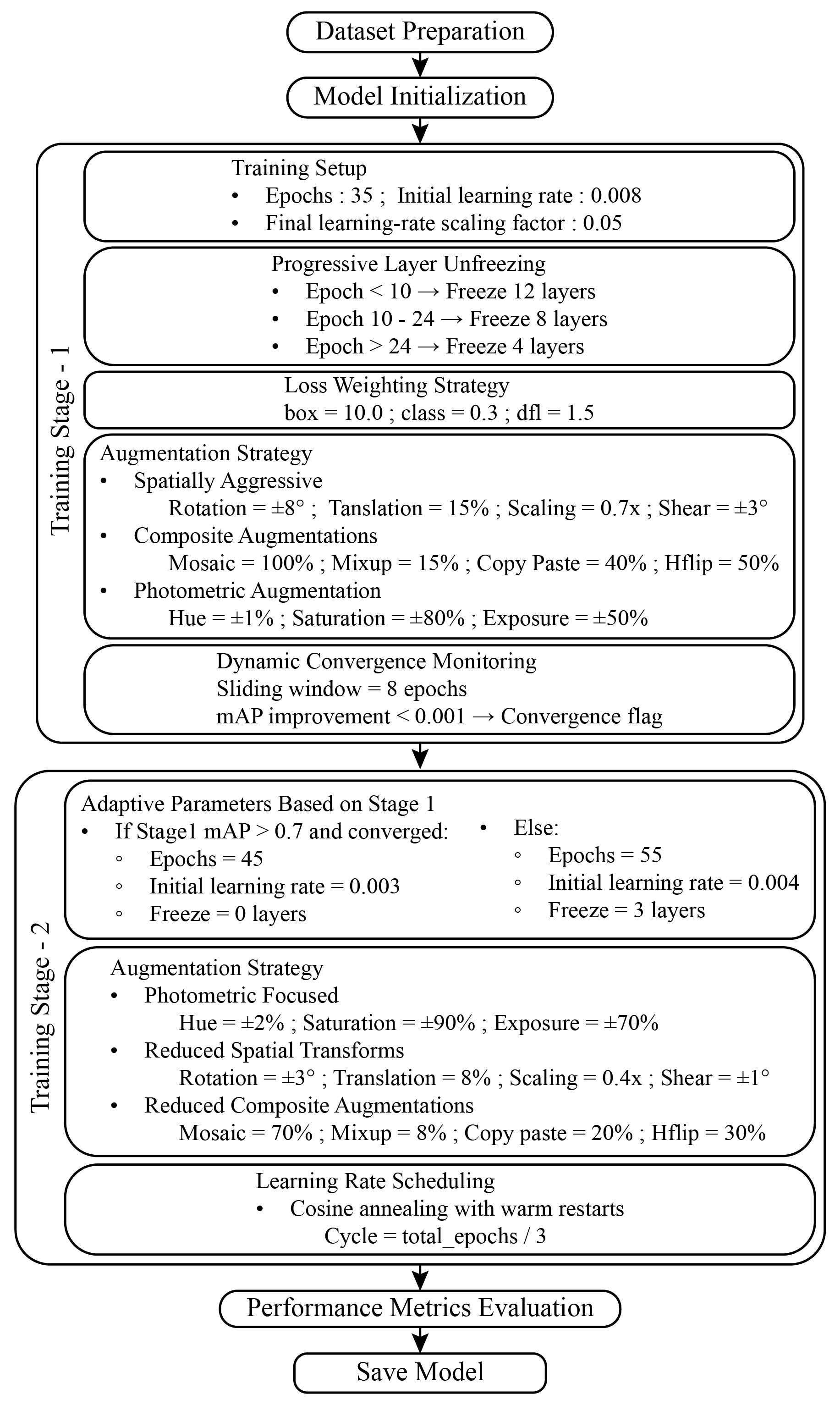}
    \caption{Proposed Algorithm for Localization Model}
    \label{fig:flowchart}
\end{figure}

The second stage implemented the previous stage’s performance-adaptive fine-tuning. If convergence was achieved (mAP > 0.7) in Stage 1, it used 45 epochs with complete unfreezing and conservative learning; otherwise, it extended to 55 epochs with light freezing and moderate learning. Augmentation in this stage was shifted toward photometric robustness with increased HSV variations (±2\% hue, ±90\% saturation, ±70\% brightness), while reducing spatial transformations (±3° rotation, 8\% translation, 0.4× scaling, ±1° shear). Composite augmentations were scaled down to 70\% mosaic, 8\% mix-up, 20\% copy-paste, and 30\% horizontal flipping to stabilize the fine-tuning. This stage employed cosine-annealed learning-rate cycles to improve generalization and reduce false positives. A dynamic convergence-checking system also monitored the model’s improvement and automatically adjusted the duration and hyperparameters of this stage. This helped to avoid overtraining and improve reliability. The complete algorithm of the localization model is presented in the flowchart in Figure \ref{fig:flowchart}.

The proposed algorithm ensures a robust and reliable detection model specifically developed for Bangla license plate detection, capable of handling diverse real-world conditions like variable lighting, angles, occlusions, and camera quality. It uses aggressive spatial transformations initially for viewpoint robustness, then emphasises photometric variations for lighting-condition adaptability. Mixup and copy-paste augmentations specifically address class imbalance and improve small-object detection. The adaptive scheduling ensures that augmentation intensity matches training phase needs: aggressive during feature learning and conservative during refinement.

\section{Methodology for License Plate Recognition}
To perform Optical Character Recognition (OCR) on the detected Bangla license plates, we adopted an end-to-end sequence-to-sequence framework. This approach treats OCR as a machine translation task, translating image features into a corresponding text sequence.

\subsection{Model Architecture}
We utilized the \texttt{VisionEncoderDecoderModel} framework from the Hugging Face Transformers library to experiment with various pre-trained vision encoders and language decoders. We evaluated the following architectures to identify the most effective combination for Bangla license plate recognition:

\begin{itemize}
    \item \textbf{ViT + BanglaBERT:} We explored a language-specific hybrid model using the same ViT encoder paired with \texttt{sagorsarker/bangla-bert-base}, a BERT-based model \cite{Sagor_2020} pre-trained specifically on Bengali text.
    \item \textbf{ViT + mBART:} This configuration combines \texttt{google/vit-base-patch16-384} (Vision Transformer) as the encoder for extracting visual features and \texttt{facebook/mbart-large-50-many-to-many-mmt} as the decoder for generating text sequences.
    \item \textbf{TrOCR:} We also evaluated the \texttt{microsoft/trocr-base-stage1} model, which is an end-to-end Transformer-based OCR model pre-trained on large-scale synthetic data.
\end{itemize}

\subsection{Model Configuration}
For the generation process, we explicitly configured the model's hyperparameters to optimize for character sequence prediction and regularization. We utilized beam search for decoding and applied specific dropout rates to both the encoder and decoder to prevent overfitting. The complete configuration of the hyperparameters used is presented in Table \ref{tab:model_config}.

\begin{table}[h]
\centering
\caption{Model Configuration and Hyperparameters}
\label{tab:model_config}
\begin{tabular}{|l|c|}
\hline
\textbf{Hyperparameter} & \textbf{Value} \\ \hline
Decoder Mode (\texttt{is\_decoder}) & True \\
Add Cross Attention & True \\
Encoder Dropout Rate & 0.3 \\
Decoder Dropout Rate & 0.3 \\
Attention Probabilities Dropout & 0.3 \\
Max Sequence Length & 20 \\
Early Stopping & True \\
No Repeat N-gram Size & 0 \\
Length Penalty & 1.0 \\
Weight decay & 0.01 \\
Number of Beams & 3 \\ \hline
\end{tabular}
\end{table}

Additionally, during training, we employed mixed-precision (\texttt{fp16 = True}) to optimize memory usage and training speed.

\subsection{Training and Evaluation}
These models were trained using a custom dataset that had labelled images of Bengali license plates. The dataset was divided into a 2,134 image training set and a 534 image validation set. Moreover, the test group will only include images reserved for a hidden test split; therefore, the performance evaluation is unbiased and fair.

The training was done with the Seq2SeqTrainer, and the per-device batch size was six. The model experienced 7800 training steps. A custom compute metrics function was used to track performance.

The monitoring and evaluation of the model were based on some major metrics determined during the validation process. These measures included the Character Error Rate (CER) which is the ratio of the number of incorrect characters in the prediction string to the ground truth (use of Hard CER in which a 1.0 value would imply an exact match); the Word Error Rate (WER) which is a measure of the error rate per word; the Levenshtein Distance, which is a measure of the minimum number of one character changes needed to transform the predicted text to the ground truth; and the last are the Training Loss and Validation Loss, measures that is followed during the training procedure to check how the model is converging

\subsection{Inference and Repetition Control}
When generating text in inference, we used a beam search decoding method and a beam width of 3 (\texttt{num\_beams = 3}). In our last setup, \texttt{no\_repeat\_ngram\_size} is adjusted to 0 meaning that n-gram repetition blocking is disabled but the model may repeat sequence of characters where it is required. This option is crucial to Bangladeshi license plates, where the numeric part may be included in the six-digit IG and have any concatenation of digits, even with repeated characters and repeated short n-grams (e.g., \textbn{'১১', '১১-১১'}, or other repetitions). Enforcing an n-gram constraint (e.g., \texttt{no\_repeat\_ngram\_size = 2} or \texttt{3}) could incorrectly suppress valid plate numbers and reduce recognition accuracy by penalizing legitimate repetitions. Therefore, allowing unrestricted repetition provides better fidelity to real-world plate formats and avoids introducing decoding-time bias against valid digit sequences.
 \cite{holtzman2019curious}.

\section{Results}
\subsection{Performance Evaluation of the Localization Models}

The accuracy of a localization model depends on how precisely it can predict the bounding box around the target object, in this case, the license plate. Therefore, the Intersection over Union (IoU) metric plays an important role in determining prediction accuracy. This study compares the performance of different models based on a binary accuracy metric: a prediction is assigned a value of 1 if IoU > 0.7 for an image, and 0 if IoU < 0.7. Each model is evaluated on approximately 553 test samples, and the average results are presented in the Table \ref{tab:det_results} below

\begin{table}[h] 
\centering 
\caption{Evaluation Metrics of Localization Models} \label{tab:det_results} 
\resizebox{\columnwidth}{!}{
\begin{tabular}{lcccccc} 
\toprule \textbf{Model} & \textbf{Accuracy(\%)} & \textbf{Precision(\%)} & \textbf{Recall(\%)} & \textbf{F1 Score(\%)} & \textbf{IoU(\%)}\\ 
\midrule 
U-net & 91.43 & 93.19 & 89.71 & 91.41 & 82.1\\ 
YOLOv5m & 90.97 & 99.89 & 87.03 & 93.06 & 83.9\\ 
YOLOv7m & 67.15 & 95.28 & 54.59 & 69.42 & 63.5\\ 
YOLOv8m & 97.11 & \textbf{99.93} & 95.68 & 97.79 & 89.7\\ 
\makecell[l]{\textbf{YOLOv8m +} \\\textbf{Multi Stage} \\\textbf{Learning}} 
& \textbf{97.83} & 99.44 & \textbf{96.31} & \textbf{98.37} & \textbf{91.3}\\
YOLOv9m & 95.67 & 99.44 & 95.20 & 97.24 & 88.8\\ 
YOLOv11m & 96.39 & 99.87 & 95.14 & 97.51 & 89.2\\ 
\bottomrule 
\end{tabular} 
}
\end{table}

The table highlights that YOLOv8m architecture combined with a Multi-Stage Progressive Learning model outperformed the other algorithms in terms of accuracy and IoU. The inference time of this model is also satisfactory as compared with others. 

\begin{figure}[t]
    \centering
    \begin{subfigure}[b]{0.48\linewidth}
        \centering
        \includegraphics[width=\linewidth]{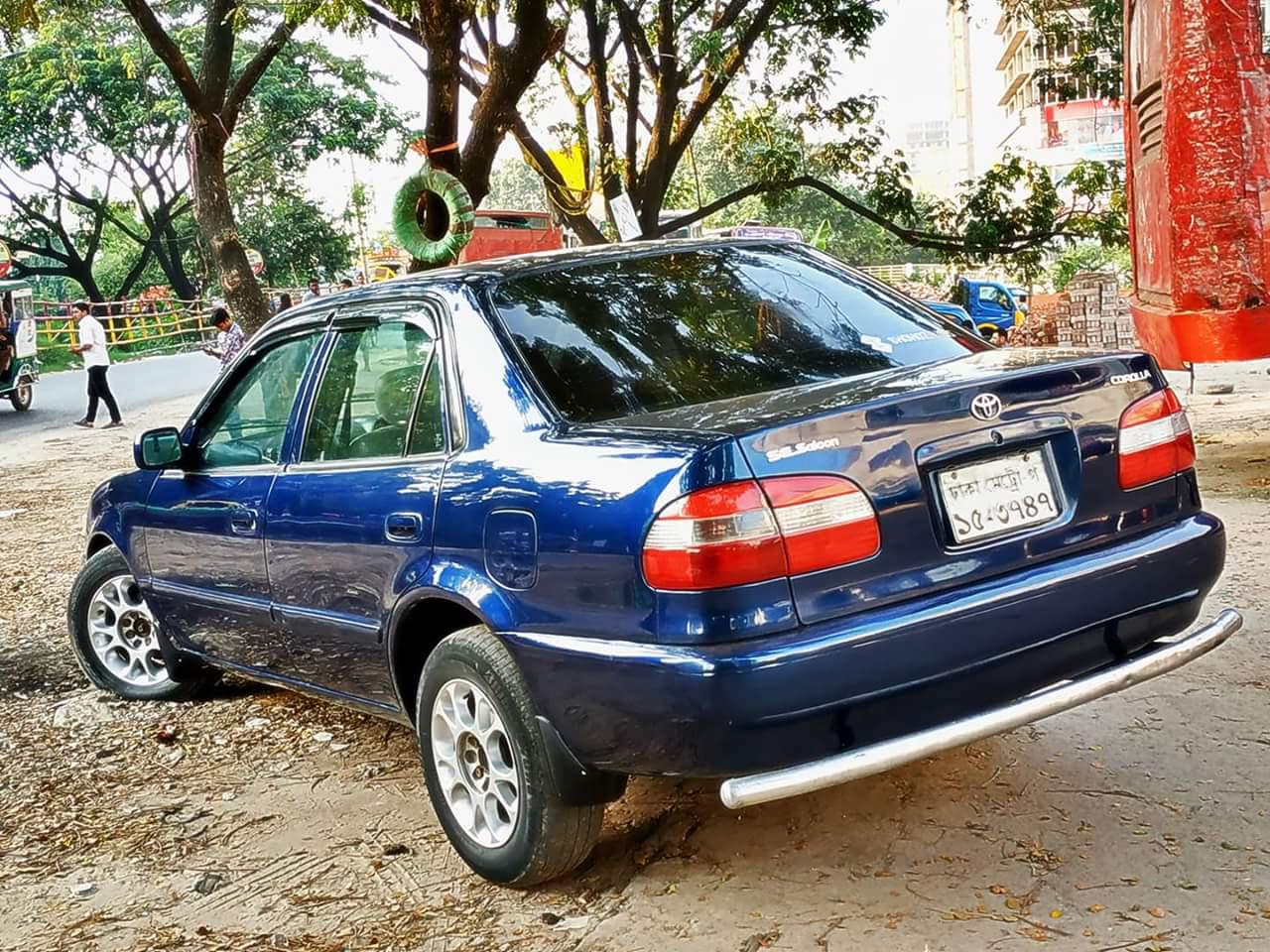}
        \caption{Test Image}
        \label{fig:image1}
    \end{subfigure}
    \hfill
    \begin{subfigure}[b]{0.48\linewidth}
        \centering
        \includegraphics[width=\linewidth]{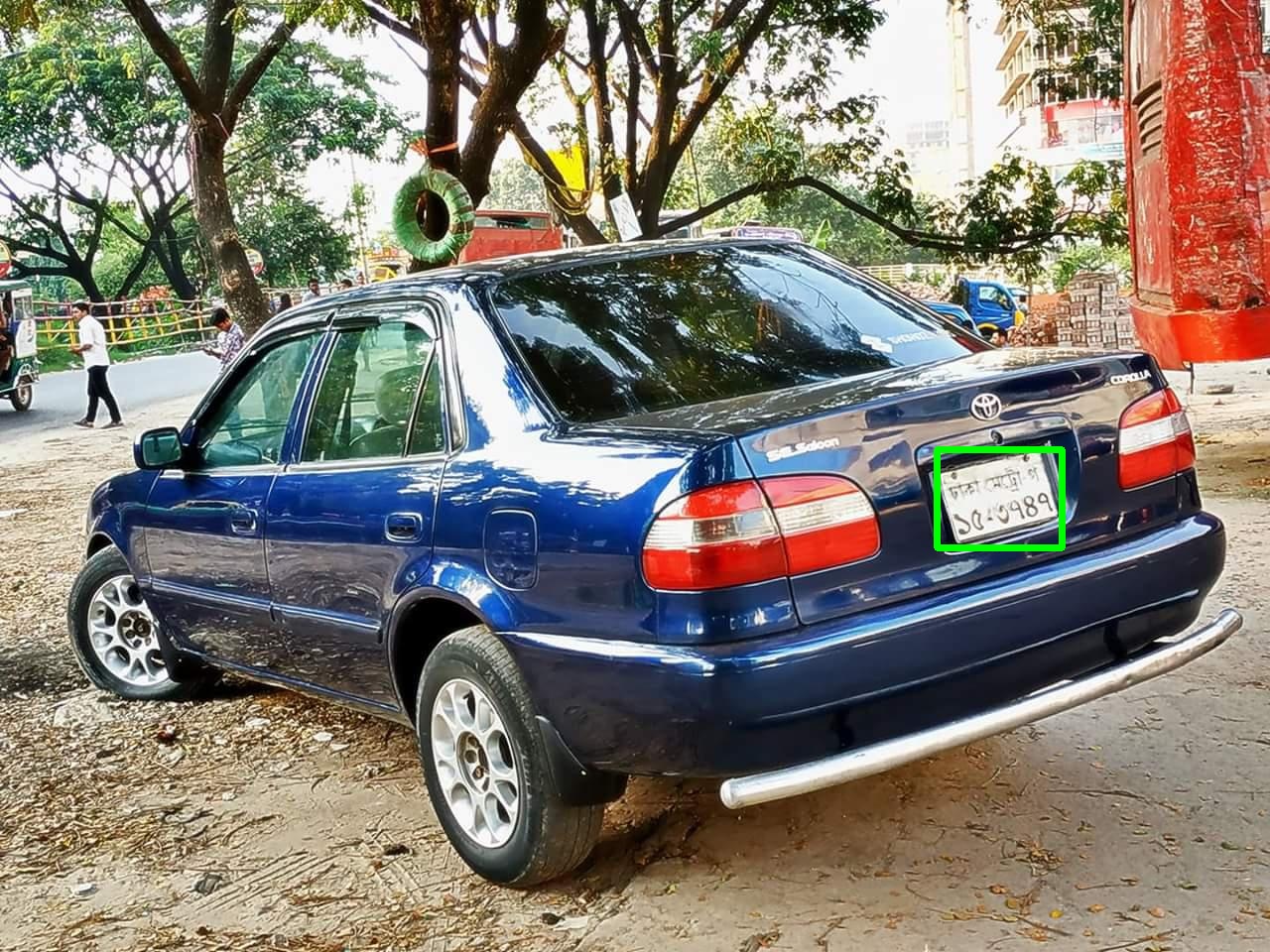}
        \caption{Detection Result}
        \label{fig:image2}
    \end{subfigure}

    \caption{License plate detection}
    \label{fig:detection1}
\end{figure}

Though image segmentation algorithms like U-Net typically perform better for objects captured at skewed or oblique camera angles, their effectiveness depends on detailed pixel-level or polygonal annotations. However, the dataset used in this study provides only axis-aligned rectangular bounding boxes, as commonly used in YOLO-based detectors. This limits the utilization of the U-net’s segmentation capability, resulting in lower performance compared to YOLO. More precise polygonal or rotated-box annotations would be necessary to fully utilize the advantages of segmentation-based models over bounding-box–based architectures. A sample detection with the localization model is shown in Figure \ref{fig:detection1}.

\subsection{Performance of OCR Models}
We present the quantitative evaluation of the three experimental OCR architectures on the Bangla license plate validation dataset. We primarily focus on CER and WER to assess the transcription accuracy of each encoder-decoder configuration. Table \ref{tab:model_performance} summarizes the comparative results, identifying the most effective model for this specific domain. Table \ref{tab:ocr_results} presents the best evaluation metrics achieved by the model during the training process.

\vspace{4pt}

\begin{table}[h!]
\centering
\caption{Performance Comparison of OCR Models on external dataset}
\label{tab:model_performance}
\resizebox{\columnwidth}{!}{
\begin{tabular}{lccccc}
\midrule 
\textbf{Model} & \textbf{Encoder} & \textbf{Decoder} & \textbf{CER} $\downarrow$ & \textbf{WER} $\downarrow$ \\ 
\midrule 
\textbf{ViT + B-BERT} & ViT-Base & BanglaBERT & \textbf{0.1323} & 0.1068 \\ \midrule 
TrOCR & TrOCR-Base & TrOCR-Base & 0.1363 & 0.1583 \\ \midrule 
ViT + mBART & ViT-Base & mBART-50 & 0.4438 & \textbf{0.0935} \\ \hline
\end{tabular}
}
\end{table}

\begin{table}[h]
\centering
\caption{Summary of Best OCR Model Performance}
\label{tab:ocr_results}
\begin{tabular}{lcc}
\toprule
\textbf{Metric} & \textbf{Best Value} \\
\midrule
Validation Loss & 0.4101 \\
Character Error Rate (CER) & 0.1323 \\
Word Error Rate (WER) & 0.1068 \\
Levenshtein Distance & 3.02 \\
\bottomrule
\end{tabular}
\end{table}

\noindent \textbf{Analysis of Training Dynamics:}
The training progression of the three evaluated architectures (ViT+mBART, ViT+BanglaBERT, and TrOCR) was monitored over 7,800 steps. The following subsections analyze the convergence trends for each metric individually.

\begin{figure}
    \centering
    \includegraphics[width=0.8\linewidth]{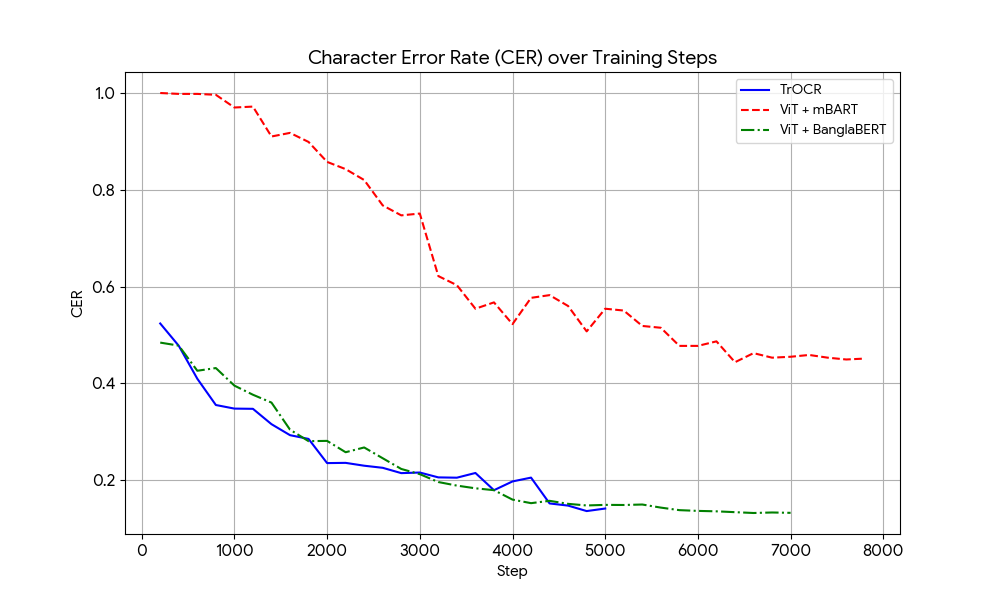}
    \caption{Comparison of Character Error Rate (CER) over training steps.}
    \label{fig:cer}
\end{figure}

\noindent \textbf{CER Analysis:}
Figure \ref{fig:cer} illustrates the decline in CER serving as the primary metric for OCR task. \texttt{ViT+BanglaBERT} (green) and \texttt{TrOCR} (blue) models both show a sharp drop at first, bringing the error rate below 0.20 in the first 2,000 steps. Efficient feature alignment between the visual encoder and the text decoder is seen in this case. The \texttt{ViT+BanglaBERT} model achieves the greatest stability, gaining a minimum CER of 0.1323. Though \texttt{TrOCR} closely follows it, it shows slightly higher variance in the later stages. In contrast, the \texttt{ViT+mBART}(red) plateaus at a significantly higher CER ($\approx 0.44$) which signifies it struggles with fine-grained characters compared to the other architectures.

\begin{figure}
    \centering
    \includegraphics[width=0.8\linewidth]{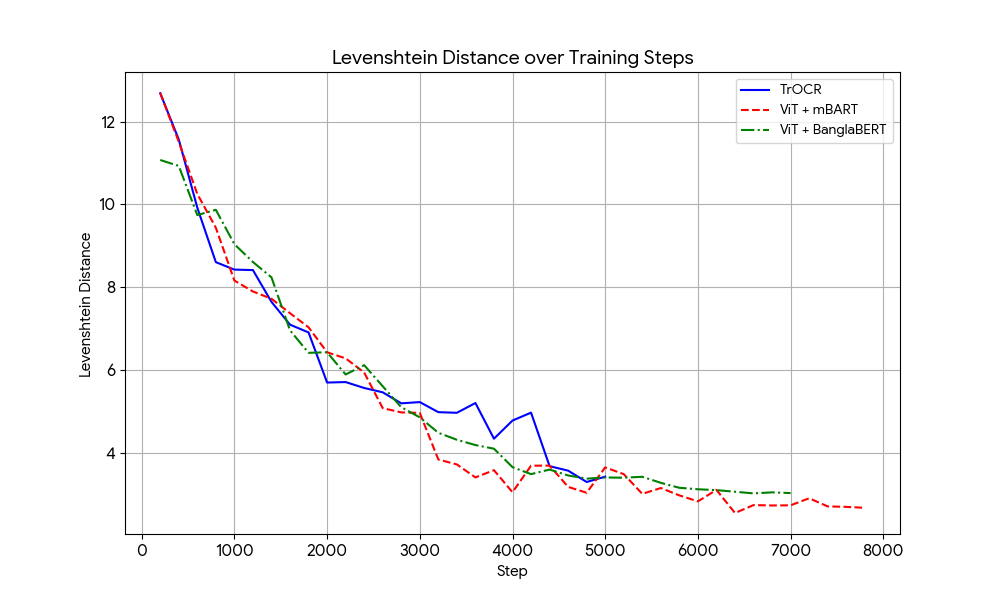}
    \caption{Comparison of Levenshtein Distance convergence.}
    \label{fig:levenshtein}
\end{figure}

\noindent \textbf{Levenshtein Distance Analysis:}
Figure \ref{fig:levenshtein} shows that the edit distance metric follows the same pattern as the CER. The \texttt{ViT+BanglaBERT} model needs the fewest single-character changes (around 3.02) to match the ground truth, which shows that it has a better understanding of characters. \texttt{TrOCR} is quite close behind, with an average distance of 3.30. On the other hand \texttt{ViT+mBART}  is behind and needs a lot more corrections for each sequence.

\begin{figure}
    \centering
    \includegraphics[width=0.8\linewidth]{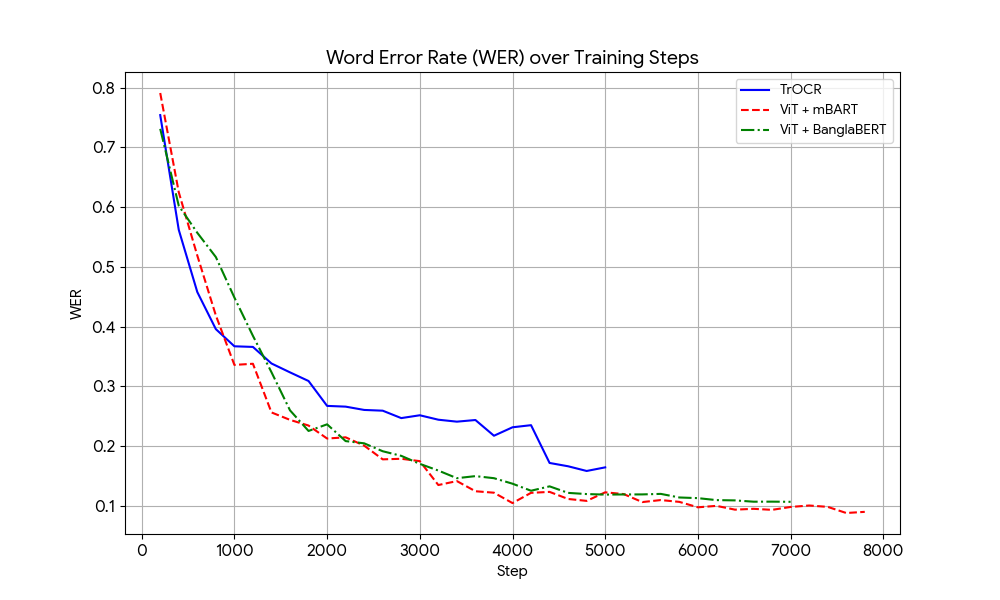}
    \caption{Comparison of Word Error Rate (WER) across models.}
    \label{fig:wer}
\end{figure}

\begin{figure}
    \centering
    \begin{subfigure}[b]{0.48\linewidth}
        \centering
        \includegraphics[width=\linewidth]{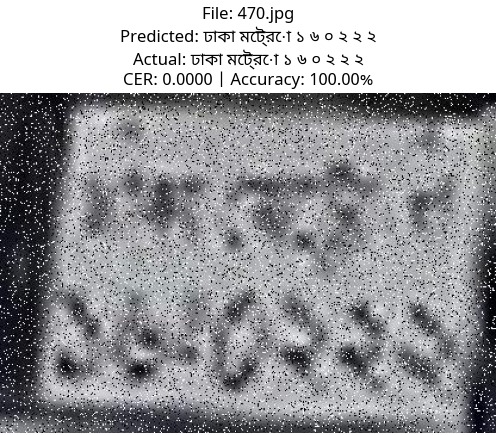}
        \caption{Correct OCR example}
        \label{fig:correct}
    \end{subfigure}
    \hfill
    \begin{subfigure}[b]{0.48\linewidth}
        \centering
        \includegraphics[width=\linewidth]{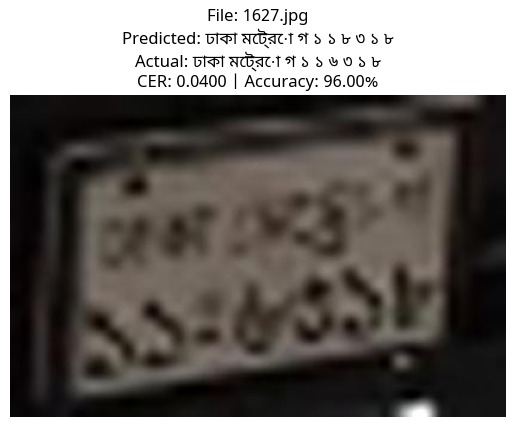}
        \caption{Incorrect OCR example}
        \label{fig:incorrect}
    \end{subfigure}
    \caption{ OCR outputs of the model using the external dataset.}
    \label{fig:ocr_examples}
\end{figure}

\noindent \textbf{WER Analysis:}
Figure \ref{fig:wer} reveals an interesting anomaly. The \texttt{ViT+mBART} model achieves a remarkably low WER (0.0935), often surpassing the \texttt{ViT+BanglaBERT} model (0.1068). This high CER but low WER suggests that while \texttt{mBART} effectively captures word boundaries and general semantics (tokenizing full words correctly). It fails to resolve specific intra word characters, which heavily penalizes the CER and Levenshtein metrics but leaves the word-level token matches intact. Figure \ref{fig:correct} and \ref{fig:incorrect} illustrates two example OCRs of the model using the external dataset.

\subsection{External Validation for the Localization Models }
To examine the performance of the model in a real-world application, the models are tested using another external dataset. The performance of the model with these images is illustrated below in Table \ref{tab:det_results2}.

\begin{table}
\centering 
\caption{Performance with External Dataset} \label{tab:det_results2} 
\resizebox{\columnwidth}{!}{
\begin{tabular}{lccccccc} 
\toprule \textbf{Model} & \textbf{Accuracy(\%)} & \textbf{Precision(\%)} & \textbf{Recall(\%)} & \textbf{F1 Score(\%)} & \textbf{IoU(\%)} \\ 
\midrule 
U-net & 80.43 & 92.10 & 82.59 & 80.79 & 60.4\\ 
YOLOv5m & 52.53 & 94.26 & 63.20 & 72.81 & 45.20\\ 
YOLOv7m & 55.12 & 97.46 & 62.74 & 75.43 & 43.04\\ 
YOLOv8m & 61.11 & 99.87 & 69.13 & 79.30 & 52.51\\ 
\makecell[l]{\textbf{YOLOv8m +} \\\textbf{Multi Stage} \\\textbf{Learning}} 
& \textbf{82.9} & \textbf{99.92} & \textbf{89.14} & \textbf{92.10} & \textbf{67.5}\\
YOLOv9m & 53.62 & 99.86 & 63.68 & 75.83 & 47.42\\ 
YOLOv11m & 54.73 & 99.88 & 64.71 & 78.02 & 48.30\\ 
\bottomrule 
\end{tabular} 
}
\end{table}

Table \ref{tab:det_results2} highlights that the F1 scores of basic YOLO-based architectures drop significantly when evaluated with images under critical conditions. This indicates that these models are highly sensitive to the training dataset’s environmental conditions and may have been overfitted. The accuracy and IoU also dropped sharply. 

\begin{figure}[b]
    \centering
    \begin{subfigure}[b]{0.48\linewidth}
        \centering
        \includegraphics[width=\linewidth]{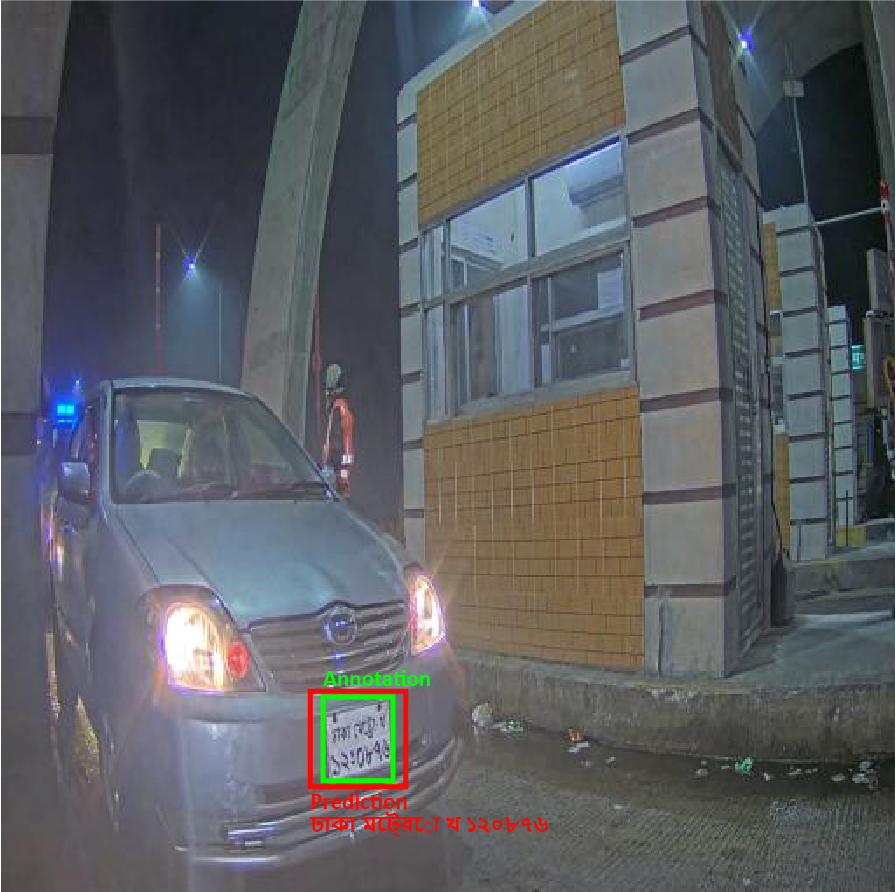}
        \label{fig:image1}
    \end{subfigure}
    \hfill
    \begin{subfigure}[b]{0.48\linewidth}
        \centering
        \includegraphics[width=\linewidth]{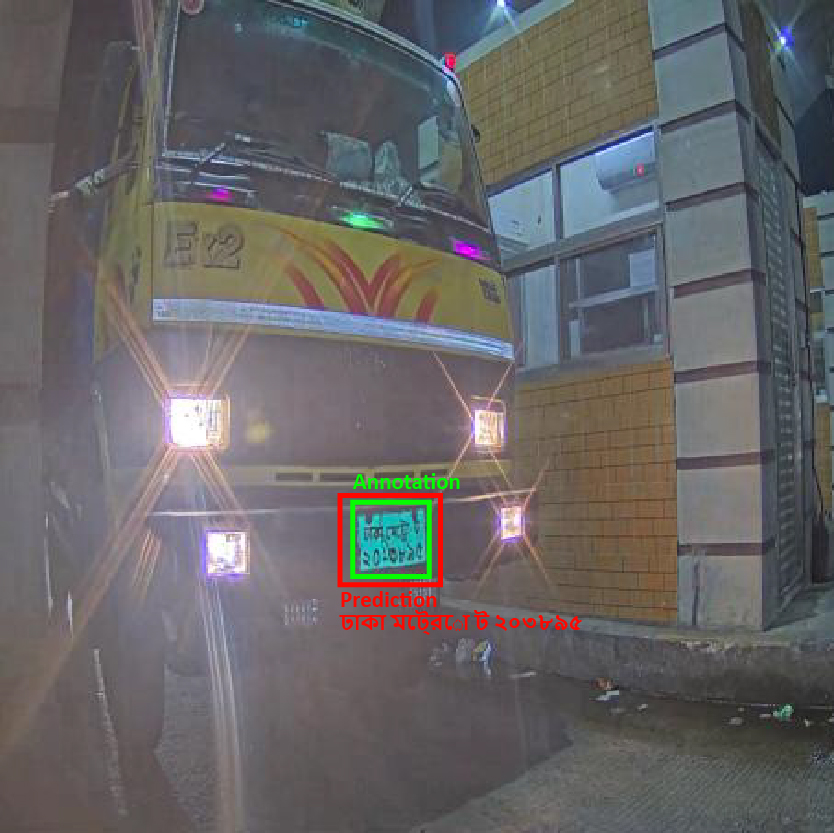}
        \label{fig:image2}
    \end{subfigure}

    \caption{Inference on External Dataset (Green = Annotation; Red = Prediction}
    \label{fig:ex_val1}
\end{figure}

On the other hand, the proposed YOLOv8m with the multi-stage progressive learning algorithm shows a comparatively consistent F1 score in both normal and critical conditions. Its accuracy and IoU also remain satisfactory in images taken under low-light or foggy environments. This consistent performance validates the model’s reliability in adverse weather or low-visibility environments, aligning well with the primary objectives of this study. Figure \ref{fig:ex_val1} illustrates two example detections of the model using the external dataset.

\begin{table}[t]
\centering 
\caption{Inference Time of Different Algorithms} \label{tab:det_results3} 
\resizebox{\columnwidth}{!}{
\begin{tabular}{lcc} 
\toprule \textbf{Model} & \textbf{With Training Dataset} & \textbf{With External Dataset} \\ 
\midrule 
U-net & 187.4 & 195.80\\ 
YOLOv5m & 101.5 & 209.43\\ 
YOLOv7m & 98.53 & 211.73\\ 
YOLOv8m & 73.9 & 89.77\\ 
\makecell[l]{\textbf{YOLOv8m +} \\\textbf{Multi Stage} \\\textbf{Learning}} 
& \textbf{83.48} & \textbf{76.50}\\
YOLOv9m & 158.51 & 208.63\\ 
YOLOv11m & 167.74 & 207.03\\ 
\bottomrule 
\end{tabular} 
}
\end{table}

To evaluate how quickly the models can successfully detect a license plate, the inference time is measured using both the test images from the training dataset and the external validation dataset. Table \ref{tab:det_results3} shows that the YOLOv8m model achieves the lowest inference time in both cases. Although applying the multi-stage progressive learning algorithm slightly increases the inference time, it results in notable improvements in both accuracy and IoU, making the trade-off worthwhile.

\section{Conclusion}
This paper introduces a comprehensive and resilient Bangla License Plate Recognition system capable of functioning reliably in challenging real-world imaging settings, including low-light operation, fluctuating noise levels and diverse camera orientations. A comparative evaluation shows that YOLOv8 enhanced with a two-stage adaptive training strategy (phase-aware augmentation, dynamic freezing and performance-based fine-tuning) delivers the most reliable outcome for localization task. ViT+BanglaBERT achieves the best character-level accuracy for optical character recognition by modeling Bengali’s linguistic and structural complexity. The results highlight that performance gains are primarily driven by the proposed training strategy rather than architectural changes alone. While YOLOv8 already provides strong baseline performance, the adaptive training significantly improves robustness under domain shifts, as evidenced by the external validation results. This suggests that training strategy design is as critical as model selection in real-world ALPR systems.

Despite these promising results, the system has certain limitations. Most of the photos used to train the model for the localization task were taken during the day time with different amounts of noise level. If the model were to utilize a more diverse dataset encompassing low-light conditions, its performance would be significantly improved. Additionally, the bounding box labels in the dataset are axis-aligned rectangular annotations. As a result, the edge-detection capabilities of pixel-wise segmentation models such as U-Net cannot be fully utilized. This may decrease the accuracy of license plate detection for the images captured at skewed orientations due to camera angles. The current system is evaluated on still images, and its scalability to continuous video streams—particularly in terms of temporal consistency and real-time processing—has not been investigated.

Future work can focus on training the model using a more comprehensive dataset covering a wider range of adverse weather conditions. Deploying the system in real-time CCTV-based traffic monitoring devices can be another important direction towards developing a fully automated traffic management system. Furthermore, enhancing computational overhead for processing real-time video feeds remains a critical domain of investigation. Addressing these issues will significantly enhance the practical applicability of the proposed system for real-world deployments such as traffic surveillance, toll collection, and law enforcement.

\section*{Acknowledgment}
All source code and curated external datasets related to this work are publicly accessible at: https://github.com/Snap0dragon/Bangla-ALPR-using-Vision-Transformer

\bibliographystyle{IEEEtran}
\bibliography{references}
\end{document}